\documentclass[11pt]{article} 
\usepackage[margin=1in]{geometry}  
\usepackage[skip=.5em]{parskip}
\usepackage{algorithm}
\usepackage{algpseudocode}
\usepackage{xcolor}
\usepackage{wrapfig}
\usepackage{bbding}
\usepackage{natbib}
\usepackage{graphicx}
\usepackage{subcaption}
\usepackage{caption}
\usepackage{booktabs}
\usepackage{amssymb}
\usepackage{color}

\usepackage{amsmath,amsfonts,bm}

\def\Figref#1{Figure~\ref{#1}}

\def\Appref#1{Appendix~\ref{#1}}

\def\Secref#1{Section~\ref{#1}}

\def\eqref#1{eqn. (\ref{#1})}

\def\Algref#1{Algorithm~\ref{#1}}

\def\Tabref#1{Table~\ref{#1}}

\def\Lineref#1{Line~\ref{#1}}

\def\1{\bm{1}}
\newcommand{\linesref}[2]{Line #1--#2}

\DeclareMathAlphabet{\mathsfit}{\encodingdefault}{\sfdefault}{m}{sl}
\SetMathAlphabet{\mathsfit}{bold}{\encodingdefault}{\sfdefault}{bx}{n}

\DeclareMathOperator*{\argmax}{arg\,max}

\newif\iffinal
\finaltrue
\iffinal
    \newcommand{\revise}[1]{}
    \newcommand{\yuxin}[1]{}
    \newcommand{\yuxinil}[1]{}
    \newcommand{\fengxue}[1]{}
    \newcommand{\diantong}[1]{}
    \newcommand{\chong}[1]{}
    \newcommand{\TBD}[1]{}
    
\else
    \newcommand{\yuxin}[1]{\todo[fancyline,color=purple!40]{YC: #1}\xspace}
    \newcommand{\yuxinil}[1]{\textbf{\textcolor{blue}{[YC: #1]}}}
    
    \newcommand{\diantong}[1]{\textcolor{magenta}{[DT]: #1}} 
    
    \newcommand{\chong}[1]{\textcolor{purple}{[Chong]: #1}} 
         
    \newcommand{\TBD}[1]{{{[{\bf TD:} \color{purple}#1]}{}}}
    \newcommand{\revise}[1]{{\textcolor{magenta}{#1}}}
\fi

\def\pfnloss{{\hat{\ell}_\theta(D\cup\{(x_i,y_i)\}_{i=1}^m)}}

\def\algname{\ensuremath{\textsc{ProfBO}}\xspace}
\def\fsbo{\textsc{FSBO}\xspace}
\def\optformer{\textsc{OptFormer}\xspace}

\def\random{\textsc{Random}\xspace}
\def\covid{\textit{Covid-B}\xspace}
\def\cancer{\textit{Cancer-B}\xspace}
\def\hpob{\textit{HPO-B}\xspace}
\def\nap{\textsc{NAP}\xspace}
\def\tnp{\textsc{TNP}\xspace}
\def\tnpp{\textsc{TNP+}\xspace}
\def\maf{\textsc{MAF}\xspace}
\def\metagp{\textsc{Meta-GP}\xspace}
\def\gp{\textsc{GP}\xspace}

\def\searchspace{\ensuremath{\mathcal{X}}}
\def\trajprior{p(\mathcal{T}^{(i)})}
\def\checkmarkG{{\Checkmark}}
\def\XG{{\XSolidBrush}}

\definecolor{darkmidnightblue}{rgb}{0,0.08,0.45}
\usepackage[colorlinks,citecolor=darkmidnightblue,urlcolor=darkmidnightblue,linkcolor=darkmidnightblue,linktocpage=true]{hyperref}
\usepackage{url}
\usepackage{xspace}
\usepackage[colorinlistoftodos]{todonotes}

\title{None To Optima in Few Shots: \\Bayesian Optimization with MDP Priors}

\author{Diantong Li \thanks{School of Data Science, The Chinese University of Hong Kong, Shenzhen, China, \texttt{diantongli@link.cuhk.edu.cn}}
\and Kyunghyun Cho \thanks{Computer Science Department, New York University, New York, NY, \texttt{kyunghyun.cho@nyu.edu}}
\and Chong Liu \thanks{Department of Computer Science, University at Albany, State University of New York, Albany, NY, \texttt{cliu24@albany.edu}}}

\date{}

\begin{document}

\maketitle
\begin{abstract}

Bayesian Optimization (BO) is an efficient tool for optimizing black-box functions, but its theoretical guarantees typically hold in the asymptotic regime. In many critical real-world applications such as drug discovery or materials design, where each evaluation can be very costly and time-consuming, BO becomes impractical for many evaluations. In this paper, we introduce the Procedure-inFormed BO (ProfBO) algorithm, which solves black-box optimization with remarkably few function evaluations. At the heart of our algorithmic design are Markov Decision Process (MDP) priors that model optimization trajectories from related source tasks, thereby capturing procedural knowledge on efficient optimization. We embed these MDP priors into a prior-fitted neural network and employ model-agnostic meta-learning for fast adaptation to new target tasks. Experiments on real-world Covid and Cancer benchmarks and hyperparameter tuning tasks demonstrate that ProfBO consistently outperforms state-of-the-art methods by achieving high-quality solutions with significantly fewer evaluations, making it ready for practical deployment.

\end{abstract}
\section{Introduction}

Bayesian Optimization (BO) is an efficient machine learning–based approach for solving global optimization problems of black-box functions where the objective functions can be highly non-convex, and the functional form or derivative is not necessarily available. Owing to its ability to optimize black-box functions, BO is particularly suited for and has been applied in many critical real-world applications such as hyperparameter optimization \citep{NIPS2012_05311655,wu2020practical}, neural architecture search \citep{kandasamy2018neural,zhou2019bayesnas}, drug discovery \citep{8539993,shields2021bayesian}, and materials design \citep{khatamsaz2023physics,tian2025materials}.

In BO settings, the objective function can only be accessed by sequential, expensive and time-consuming evaluations. At each iteration all previous evaluations guide the selection process for the next query of the objective function. To study the data efficiency of BO, numerous theoretical studies focused on establishing asymptotic regret bounds \citep{Srinivas_2012,chowdhury2017kernelized}, such as cumulative regrets and simple regrets. However, asymptotic efficiency often proves impractical in real-world experimental design scenarios, primarily due to the high cost and long turnarounds of each function evaluation. In penicillin manufacturing, for example, the pharmaceutical goal is to identify the optimal experimental control parameters (e.g., temperature, humidity, biomass concentration) that maximize yield. Unfortunately, each evaluation via the wet labs can take days or weeks and incur substantial costs, severely limiting the number of feasible evaluations per year. Notably, \cite{liang2021scalable} reported that even state-of-the-art BO algorithms require around 1,000 iterations to converge in a penicillin production simulator, equivalent to nearly 20 years if each evaluation takes one week. Also, \cite{Aldewachi2021} reported that developing a new drug cost approximately \$1.3 billion USD on average in 2018, while a failed Alzheimer's disease drug development program could take up to 5 years and cost as much as \$2.5 billion.

Therefore, to significantly accelerate the scientific discovery processes mentioned above, an urgent and important question is:
\begin{quote}
    \textit{Can we design a BO algorithm that is able to find the global optimum within very few shots, e.g., fewer than 20 evaluations?}
    
\end{quote}

\begin{wrapfigure}{r}{0.45\textwidth}
  \centering
  \includegraphics[width=0.45\textwidth]{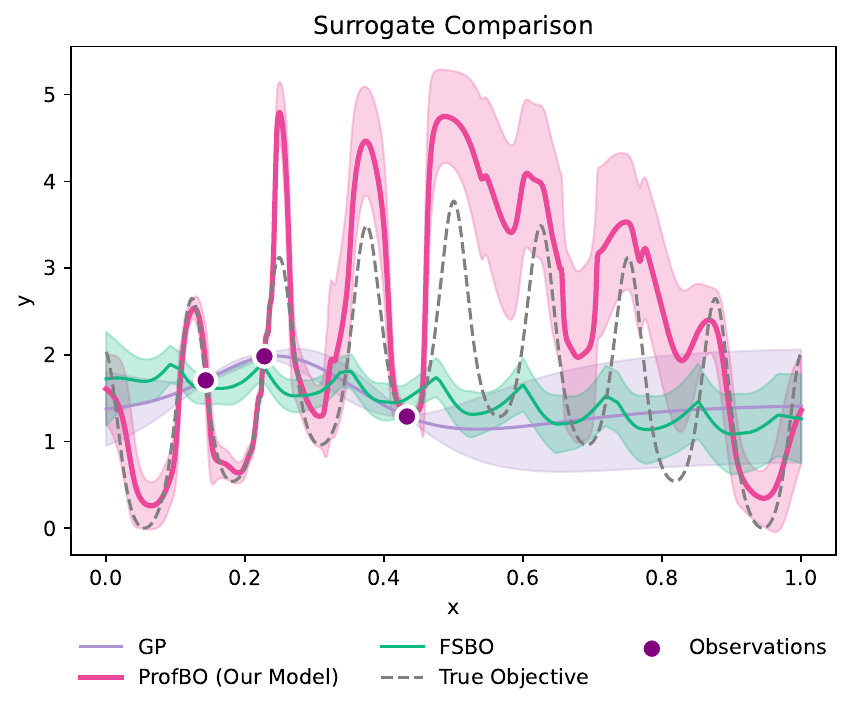}
  \caption{Comparison of function predictions with only 3 observation points. \algname models the true objective function curve significantly better than GP model and FSBO algorithm \citep{aea3f03299ff0cfea9b394f5559aa1c173f9876f}.}
 
  \label{fig: surrogate comparison}
\end{wrapfigure}

In this paper, we answer this question affirmatively by proposing a few-shot BO method, the \textbf{Pro}cedure-in\textbf{f}ormed \textbf{B}ayesian \textbf{O}ptimization (\algname). How does \algname address the problem above that initially seems impossible? The key insight behind \algname is that, although the target task permits only a few evaluations, related source tasks often have existing evaluations that can serve as rich sources of prior information. Leveraging these existing evaluations enables us to both design and accelerate a BO algorithm for the target task. In practice, these related source tasks can be the docking scores of a set of molecules evaluated on different receptors \citep{liu2023drugimprover}, or evaluations of the same supervised-learning loss function on different datasets \citep{arango2021hpoblargescalereproduciblebenchmark}.

The most straightforward way to exploit such source information is to directly predict the shape of the target objective function, as studied in \cite{aea3f03299ff0cfea9b394f5559aa1c173f9876f, RGPE, wang2024pre}. In this paper, however, we propose a framework that optimizes the target objective function by leveraging optimization trajectories of related source tasks in principled Bayesian perspective. Specifically, we use Markov Decision Process (MDP) \citep{Bellman58} priors to model source optimization trajectories, thereby capturing procedural knowledge on efficient optimization. We then embed these MDP priors into a Prior-Fitted Neural Network (PFN) \citep{d88a5ae1673f2009704186acf2890163e6ddf4ca} and employ Model-Agnostic Meta-Learning (MAML) \citep{finn2017modelagnosticmetalearningfastadaptation} for fast adaptation to new target tasks. 

The whole \algname framework works efficiently even with very few evaluations. See Figure \ref{fig: surrogate comparison} for an example showing that with only 3 evaluations, our \algname framework models the true objective function significantly better than the standard Gaussian Process (GP) model and the FSBO algorithm \citep{aea3f03299ff0cfea9b394f5559aa1c173f9876f}.

\textbf{Contributions.} Our contributions are summarized as follows:
\begin{enumerate}
    \item With very few shots, our \algname framework can accurately and efficiently identify global optima of black-box functions. Its modular design allows easy adaptation to various configurations and input types by retraining the MDP priors part only. Both make \algname highly practical for a wide range of scientific and engineering applications.
\item While optimization trajectory information from source tasks is needed, the Bayesian framework underlying \algname \emph{eliminates} all manual steps in prior design and posterior inference via the universal PFN inference.
\item The core design of the \algname algorithm lies in its use of MDP priors, which model optimization trajectories from source tasks, thereby capturing procedural knowledge of efficient optimization. These MDP priors are then embedded into a PFN, and MAML is employed for rapid adaptation to new target tasks.
\item We establish new real-world Covid and Cancer benchmarks for few-shot BO problems and show that \algname achieves better performances than all existing state-of-the-art baselines.
\end{enumerate}

\section{Related Works}
\label{sec: related works}
\paragraph{Few-Shot Bayesian Optimization.}  
In few-shot BO, people often leverage the knowledge in the evaluations of related tasks to improve the performance of BO. One common approach is meta-surrogate design, whose goal is to transfer-learn a BO surrogate that captures the shared features of multiple tasks' response surfaces. The meta-surrogates can be synthesized from surrogates of related tasks. \cite{RGPE, e91e3be38b5da3f070a2514def1ceacce450b46c, c7808a6c5d755e96100c014cba1b067de1fc1c1f} used an ensemble surrogate with weights based on metrics like task similarity or model uncertainty from related task evaluations. Meta-prior design also induces the corresponding meta-surrogate, where knowledge from related tasks is transferred to the prior distribution of the target task. \cite{NIPS2013_f33ba15e, poloczek2016warmstartingbayesianoptimization, pmlr-v33-yogatama14, NEURIPS2019_6754e06e, pmlr-v151-tighineanu22a, aea3f03299ff0cfea9b394f5559aa1c173f9876f} proposed to design novel kernels for GP functional priors. \cite{wang2024pre} proposed to obtain a target task prior by minimizing the KL divergence between it and its related tasks. \cite{0b9f042b3588ed8588e907d7d0e139a008c517cd} designed a linear function prior for the target objective whose feature is processed by a meta-trained neural network, and the inference is facilitated by Bayesian linear regression. However, most existing work relies on manually designed ensemble weights or tractable priors, and meta-learns only the response surface, while our approach \textit{eliminates} those manual designs through the universal PFN inference, and introduces a trajectory surrogate that captures procedural information.

Acquisition function design is another important approach of few-shot BO, which also employs ensemble methods \citep{c40289911543aff6de25de0dc7a58f11942de715}, similar to the surrogate design. Some work reframed BO as sequence modeling or Reinforcement Learning (RL).
According to \cite{bai2023transferlearningbayesianoptimization}, search space and initialization design \citep{2876680f59afd12482a8c78bec6723213b097b63, f349824dbbe0aba95dc1581c81f8ba0b210fea52, d7bb5f8bbd0a2527d97fd4829c2a9edd03491878, 9be7e7579fbec5d45e3e6ea1c4465258225a183d, ca74e160b47c669f889152935f27a72b4c205944} also provide important information for few-shot BO, 
and their methods primarily leverage source-domain information to provide a warm start for the target task or narrow down the search space, which is far from the techniques used in our work.

\textbf{Meta-Optimization with Sequence Modeling.} 
An emerging trend in leveraging meta-data in optimization is to frame optimization as a generic sequential decision process, thus the knowledge transfer is not limited to BO components. Those methods have demonstrated state-of-the-art results in various hyperparameter optimization and drug discovery problems.
The meta-acquisition learning method \citep{774f5e2494f037302c58b7fa549c4f1cabf7295e, hsieh2021reinforcedfewshotacquisitionfunction} reframes BO (or optimization) as a MDP. Optimization trajectories are generated during their own training processes. They meta-train a RL agent on evaluations of related tasks and conduct optimization on target tasks. \cite{iwata2021endtoendlearningdeepkernel, maraval2023endtoend} trained the acquisition function and surrogate model end-to-end, defining the state space solely as historical evaluations, which is consistent with this work. Unlike these approaches, we employ a lightweight RL agent to extract procedural experience, and generate optimization trajectories separately from the training process. And our agent is not used to optimize the target tasks, but serves as the prior distribution of the optimization trajectory.

The Transformer \citep{vaswani2023attentionneed} architecture excels in sequence prediction and in-context learning. Transformer-based BO surrogate learning methods include Transformer Neural Process (\tnp) \citep{8ef3f06353f65ca653b396b70d089f793bfc168b, b71e6247b14f62f9fb134e3aedff18b3cb082cb9}, and leveraging the knowledge in Large Language Models (LLMs) to conduct in-context learning \citep{ramos2025bayesianoptimizationcatalysisincontext, 3d3f0d7eb69f063e9539dd45df971f007b094734}. \cite{943cd7cf7ba21911561d03228dc2dd3f397168c9, embed_then_regress} framed optimization as a sequence prediction task, using LLMs to jointly predict the next query and response, leveraging evaluations of other optimization algorithms.  Our approach also uses Transformer to facilitate in-context learning, but adopts a modular design that allows easy adaptation by retraining only the MDP
priors for various configurations and input types.

\section{Preliminaries}\label{sec: prelim}

\subsection{Problem Statement}
Let $[n]$ denote the set $\{1,\cdots, n\}, n\in \mathbb{N^+}$. We consider the target problem of finding the global optimum of a black-box function $f: \mathcal{X} \rightarrow \mathbb{R}$:
\begin{align}
    x^* = \arg\max_{x\in \searchspace}f(x),
    \label{eq:problem}
\end{align}
where $\mathcal{X} \subseteq \mathbb{R}^d$ is the function domain and $d$ is the input dimension. $f$ is said to be a black-box function because its closed-form expression or the derivative is not necessarily known, and it can be a non-linear non-convex function. We learn from $f$ only through sequential noisy evaluations. Throughout $T$ iterations, at each iteration $t \in [T]$, its evaluation is given by $y_t = f(x_t) + \zeta_t$ where $\zeta_t$ is the noise. Our goal is to solve this problem within a few shots, e.g., $T \leq 20$.

Given the limited information gained from each evaluation, we assume access to historical source knowledge that can accelerate the target optimization process, enabling completion within only a few evaluations. Formally, historical knowledge can be $D^{(i)} = \{x_\tau^{(i)},y_\tau^{(i)}\}_{\tau = 1}^{n_i}, \forall i \in [N]$ generated by $N$ source black-box functions $f^{(1)},\cdots,f^{(N)}: \mathcal{X} \rightarrow \mathbb{R}$ where $n_i$ denotes the evaluation length of $f^{(i)}$. 
Additionally, we define $\trajprior$ as the process of sampling from $f^{(i)}$, which can also be comprehended as the distribution of optimization trajectories of $f^{(i)}$ under some policy.
$p(\cdot)$ or $p(\cdot|\cdot)$ denotes a (conditional) probability density function or its corresponding distribution. If $p$ is a distribution of a function $f:\mathcal{X}\rightarrow \mathbb{R}$, it will specify the distribution of $f(x)$ with probability density function $p(\cdot|x)$ where $x \in \searchspace$.

\subsection{Background}

\textbf{Bayesian Optimization (BO).} To solve \eqref{eq:problem}, BO usually assumes $f$ is drawn from a functional prior, e.g., a GP prior. 
At iteration $t$, conditioning on historical evaluations $D_{t-1}=\{x_\tau,y_\tau\}_{\tau=1}^{t-1}$, we denote the posterior predictive distribution as $p(\cdot|x,D_{t-1})$.
Then the next query $x_t$ is chosen by $x_t =\argmax_{x \in \mathcal{X}} \alpha_t(x)$ where $\alpha_t$ is the acquisition function built on $p(\cdot|x, D_{t-1})$. Upper Confidence Bound (UCB) \citep{Srinivas_2012}, Expected Improvement (EI) \citep{10.1023/A:1008306431147}, Thompson sampling \citep{thompson1933likelihood}, and knowledge gradient \citep{frazier2008knowledge} are commonly used as acquisition functions in practice.

\textbf{Prior-Fitted Neural Networks (PFNs).} The motivation of applying PFNs \citep{d88a5ae1673f2009704186acf2890163e6ddf4ca} in BO is to leverage the meta-trained Transformer architecture to perform principled Bayesian inference in a single forward pass, enabling superior few-shot learning and accurate predictions with interpretable uncertainty quantification. It outperforms traditional GP (see \Figref{fig: surrogate comparison}), offering greater flexibility in the choice of prior and faster inference by avoiding the matrix inversion in GP posterior inference, thereby enabling our proposed MDP priors (\Secref{sec:mdp_prior}). Given any observation set $D$ and an inference location $x\in \searchspace$, the PFN $q_\theta(\cdot|x,D)$ approximates $p(\cdot|x,D)$ through a bar distribution. In practice, we evaluate a sequence of queries $\{x_i\}_{i=1}^m$ in parallel. \Figref{fig: PFNs architecture} illustrates the forward pass and attention mask with three contextual observations and two inference locations. Our settings of PFN's regression head and network structure are shown in \Appref{app: PFN setting}. 

\textbf{Markov Decision Process (MDP).} An MDP \citep{Sutton1998} is typically defined as a 5-tuple $\mathcal{M} = \{\mathcal{S}, \mathcal{A}, \mathcal{P}, \mathcal{R}, \gamma \}$ where $\mathcal{S}$ is the state space, $\mathcal{A}$ is the action space, $\mathcal{P}$ is the transition probability space and $p(s_{t+1}|s_t,a_t) \in \mathcal{P}$ is the transition probability from state $s_{t} \in \mathcal{S}$ to state $s_{t+1} \in \mathcal{S}$ if the agent takes action $a_t \in \mathcal{A}$ at time $t$, $\mathcal{R} \subseteq \mathbb{R}$ is the reward space and $r_t(s_t,a_t) \in \mathcal{R}$ is the reward of taking action $a_t$ at state $s_t$, and $\gamma \in (0,1]$ is a discount factor. A classical RL problem is to train an agent by finding the optimal policy $\pi^*$ that maximizes the expected cumulative discounted reward $\pi^* = \arg\max_{\pi}\mathbb{E}_{}\left[\sum_{t=1}^\infty\gamma^{t-1}r_t\right]$ where the expectation is taken w.r.t. MDP.

\textbf{Model-Agnostic Meta-Learning (MAML).} MAML \citep{finn2017modelagnosticmetalearningfastadaptation} is a gradient-based meta-learning framework designed to train a model \(M_{\theta}\) that can rapidly adapt across a collection of tasks \(\{\mathcal{T}^{(i)}\}_{i=1}^N\). Each task \(\mathcal{T}^{(i)}\) specifies a loss \(L^{(i)}(\theta)\). The meta-objective minimizes the meta loss defined as $L_\mathrm{meta}(\theta):= \sum_{i=1}^N L^{(i)}(\theta - \beta \cdot\partial_{\theta}L^{(i)}(\theta))$
where $\beta$ is an inner step size. By descending with respect to the meta loss, the model learns the common internal representation across $N$ tasks.

\section{The ProfBO Algorithm} \label{sec: method}
\begin{figure}[t] 
    \centering
    \begin{minipage}{0.99\linewidth}
    \includegraphics[width=\textwidth]{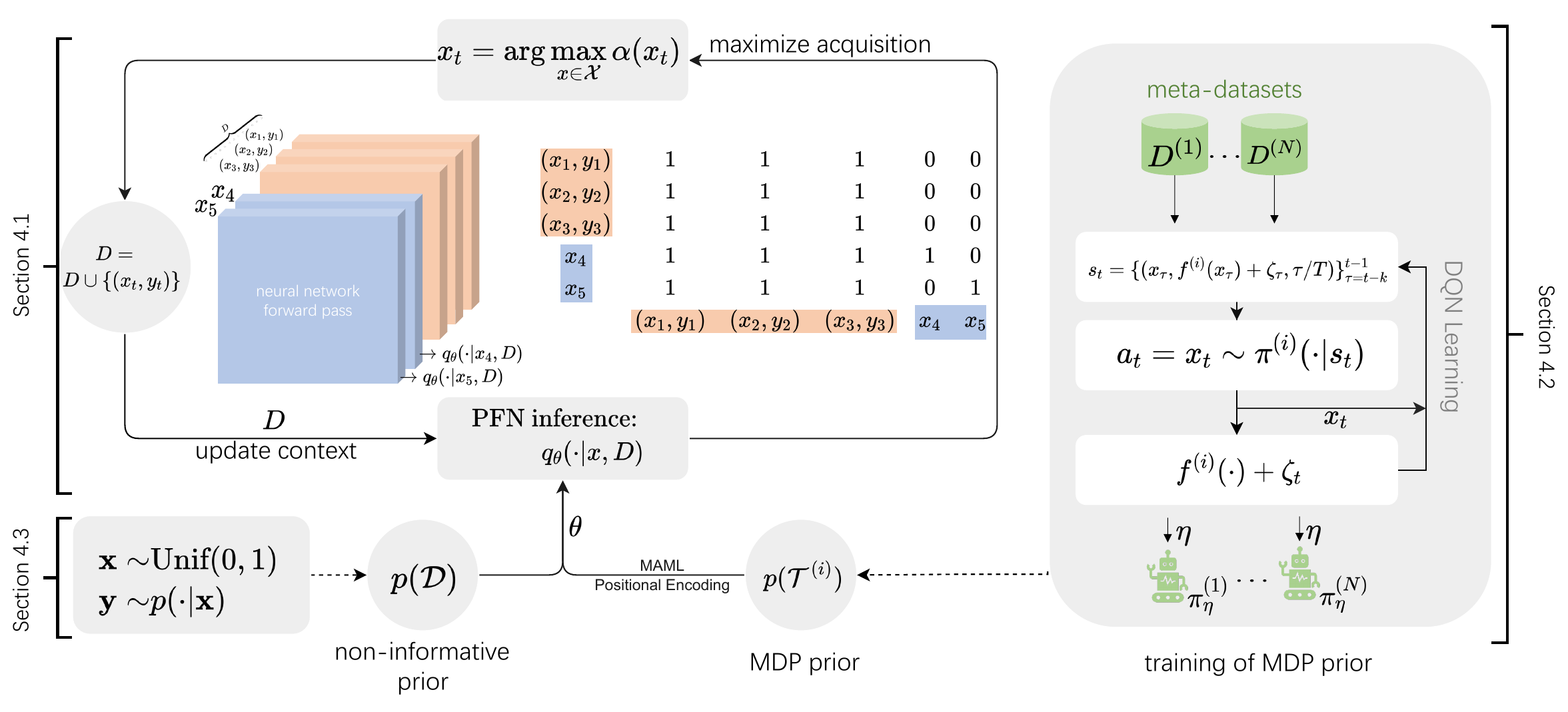}
    \end{minipage}
    \caption{Overview of our \algname framework. [Left] The BO loop: a PFN \citep{d88a5ae1673f2009704186acf2890163e6ddf4ca} model pre-trained with a non-informative prior (e.g., GP) and fine-tuned with MDP priors from source tasks. Fine-tuning uses positional encoding and MAML \citep{finn2017modelagnosticmetalearningfastadaptation} for better knowledge transfer. The PFN performs posterior inference on context $D$ via Transformer attention, with outputs interpreted as logits of a bar distribution. [Right] MDP prior training: for each meta-dataset $D^{(i)}$, a Deep Q-Network (DQN) \citep{mnih2013playingatarideepreinforcement} policy generates optimization trajectories of the corresponding objectives. Right flowchart is attributed to Figure 1 in~\cite{774f5e2494f037302c58b7fa549c4f1cabf7295e}.}
\label{fig: PFNs architecture}
\end{figure}

In this section, we show details of our \algname algorithm, a Bayesian framework that can find high-quality solutions to the black-box function optimization problem within very few shots where $T$ can be fewer than $20$. The key design of \algname lies in how it can efficiently transfer the knowledge from historical optimization trajectories obtained from related source tasks to accelerate its optimization process in target task. The core procedure is summarized in \Figref{fig: PFNs architecture}. 

The \algname framework is grounded in a principled Bayesian perspective on optimization trajectories. Under the GP assumption in standard BO, at iteration $t$, all queries in $D_{t-1}$ are treated as i.i.d. uniformly distributed, ignoring the fact that $D_{t-1}$ is actually a BO trajectory. To fully make use of this information to accelerate the BO process in the target task, we aim to construct a new surrogate model that replaces GP with a set of optimization trajectory priors, $\trajprior, \forall i \in [N]$ (Figure \ref{fig: PFNs architecture} right, Section \ref{sec:mdp_prior}). To facilitate posterior inference under such a prior, we introduce the PFN model $q_\theta(\cdot|x,D_{t-1})$ as a proxy for $p(\cdot|x,D_{t-1})$ (Figure \ref{fig: PFNs architecture} left-top, Section \ref{sec: pfn framework}). By leveraging observations from the source tasks $\{D^{(i)}\}_{i=1}^N$, MAML enables the PFN model to rapidly adapt to the unknown trajectory of the target task $f$, while preventing it from learning spurious temporal correlations (Figure \ref{fig: PFNs architecture} left-bottom, Section \ref{sec: MAML for traj surrogate}).

\subsection{The PFN Framework}\label{sec: pfn framework}

First, we introduce the PFN framework used in the BO loop where the surrogate model $p(\cdot|x,D_{t-1})$ is obtained by updating the observation set $D_{t-1}$ and conducting posterior inference. PFN conducts simulation-based inference through the forward pass of a Transformer model. It directly takes $D_{t-1}, x$ as inputs and outputs a discrete approximation of $p(\cdot|x,D_{t-1})$. Each query-evaluation pair in $D_{t-1}$ attends to each other through the Transformer attention, and are attended by the position of the query $x$. The final layer of the query network is converted to logits of $p(\cdot|x,D_{t-1})$, through which we can compute approximated acquisition functions, such as UCB and EI. 

We assume that the evaluations of $f$ is generated by the process $p(\mathcal{D})$, which is similar to $\trajprior$ defined in \Secref{sec: prelim} corresponding to a certain process sampling from $f$. To train the PFN to conduct posterior inference for $p(\mathcal{D})$, one only need to minimize the KL divergence between the PFN approximation $q_\theta(\cdot|x,D)$ and the ground-truth $p(\cdot|x,D)$, where $\theta$ is the model parameter and $D$ is any context. It is equivalent to a more tractable negative log-likelihood term $\mathbb{E}_{D \cup \{(x,y)\} \sim p(\mathcal{D})}[-\log q_{\theta}(y|x,D)]$ plus a constant, where $\{(x,y)\}\cup D$ are sampled from $p(\mathcal{D})$ \footnote{The derivation can also be found in Appendix A in \cite{d88a5ae1673f2009704186acf2890163e6ddf4ca}.}. In each gradient descent step, we aim to teach the model to make inference for context $D$ at $m$ inference locations, thus the step-wise loss is
\begin{align}
\label{eq: pfn loss}
    \pfnloss =\sum_{i=1}^{m}-\log q_{\theta}(y_i|x_i, D), \quad D\cup\{(x_i,y_i)\}_{i=1}^{m} \sim p(\mathcal{D})
\end{align}
We use a fixed total sequence length for each training step, and $m$ in each step is chosen to randomly split the sequence.
The flexible choice of $p(\mathcal{D})$ provides the possibility to infer any complicated while easy-to-generate prior. Thus, we construct simulators of $\trajprior$ based on meta-datasets of related tasks $D^{(i)}, \forall i \in [N]$ (\Secref{sec:mdp_prior}) and train the PFN with the supervised loss in \eqref{eq: pfn loss}, yielding a desired trajectory surrogate for BO.

\subsection{Modeling Optimization Trajectory with MDP}\label{sec:mdp_prior}
We propose a novel prior for Bayesian optimization that models $\trajprior$, the prior distribution of optimization trajectories, with MDP \citep{Bellman58, Sutton1998}. In this MDP, at time $t$, the action $a_t$ is defined as the next query point $x_t \in \searchspace$. The state $s_t$ consists of the previous $k$ evaluations $\{x_{\tau}, f^{(i)}(x_{\tau})+\zeta_\tau, \tau/T\}_{\tau=t-k}^{t-1}$, and the transition is adding a new evaluation $(x_t, f^{(i)}(x_t) + \zeta_t, t/T)$
while removing the oldest evaluation obtained at time $\tau = t-k$. The reward is the negative simple regret $r_t = \max_{\tau \leq t} f^{(i)}(x_\tau) - \max_{x \in \mathcal{X}} f^{(i)}(x)$. The agent thus performs $T$-shot optimization of $f$. We train an agent for each source task $f^{(i)}$ via a DQN policy \citep{mnih2013playingatarideepreinforcement}, and denote the resulting policy parametrized by $\eta$ as $\pi_\eta^{(i)}$ (\Figref{fig: PFNs architecture} right). Trajectories sampled according to policy $\pi_\eta^{(i)}$ and state transition defined above are then used to train PFNs.

The DQN agent approximates the optimal Q-function with $Q_\eta(s,a)$ and updates $\eta$ iteratively during training on dataset $D^{(i)}$ until surpassing random search (details in \Appref{app: RL setting}). Its parametric nature allows efficient batched trajectory generation on GPUs, which is crucial for PFN training.

\subsection{Model-Agnostic Meta-Learning for Trajectory Surrogate}
\label{sec: MAML for traj surrogate}
Next, we show how \algname trains the PFN framework using MAML for trajectory surrogate, with details shown in \Algref{alg: train PFNs}. Let $\text{GD}_\theta[f]:= \theta - \beta \cdot \partial_{\theta}f$ denote the gradient descent operator where $\beta$ is a learning rate. In the pre-training stage (\linesref{\ref{line: pretrain start}}{\ref{line: pretrain stop}}), we train the PFN conventionally using synthetic samples from common function priors $p(\mathcal{D})$, such as GPs. As shown in \cite{b71e6247b14f62f9fb134e3aedff18b3cb082cb9}, PFNs trained with non-trajectory functional priors are well-suitable BO surrogates, as they generate sensible mean predictions with uncertainty quantification. Therefore, $p(\mathcal{D})$ provides a warm start for the following stage. Moreover, since trajectories may not cover the entire response surface of the objective, training with $p(\mathcal{D})$ stabilizes regions unexplored by $\trajprior$s. Each iteration involves gradient descent (\Lineref{line: update pretrain}) on the loss from a batch of samples (\Lineref{line: sample pretrain}), enabling the PFN to infer across more diverse contexts.

\begin{algorithm}[!htbp]
\caption{PFN training of \algname} \label{alg: train PFNs}
\textbf{Inputs:} Prior over general distribution $p(\mathcal{D})$; prior over source tasks $\{\trajprior\}_{i=1}^N$, numbers of iterations $K_1,K_2$, initialized PFNs parameter $\theta$.
\begin{algorithmic}[1]
\For{$j \in [K_1]$} \label{line: pretrain start}
    \State Sample $D\cup\{(x_i,y_i)\}_{i=1}^m\sim p(\mathcal{D})$ \label{line: sample pretrain}
    \State Update $\theta = \text{GD}_\theta[\pfnloss]$\label{line: update pretrain} \Comment{standard gradient descent}
\EndFor \label{line: pretrain stop}
\For{$j \in [K_2]$} \label{line: finetune start}
\State sample a batch of priors $\Pi \subset \{\trajprior\}_{i=1}^N$  \Comment{sample from MDP prior}
\For{$p \in \mathcal{P}$} 
    \State  sample $D\cup\{(x_i,y_i)\}_{i=1}^m\sim p$ \label{line: sample agent fine tune}
    \State Compute $\theta_\pi = \text{GD}_\theta[\pfnloss]$  \label{line: agent specific update} \Comment{with positional encodings}
\EndFor
\State Update $\theta = \text{GD}_\theta\left[\sum_{\pi\in \Pi} \hat{\ell}_{\theta_\pi}(D\cup\{(x_i,y_i)\}_{i=1}^m)\right]$ \Comment{MAML update}\label{line: global update}
\EndFor \label{line: finetune end}
\end{algorithmic}
\textbf{Output:} {$q_\theta$}
\end{algorithm}

In the fine-tuning stage (\linesref{\ref{line: finetune start}}{\ref{line: finetune end}}), we fine-tune the PFN using samples from all $\trajprior, \forall i \in [N]$, incorporating both MAML \citep{finn2017modelagnosticmetalearningfastadaptation} and positional encoding. We introduce the positional encoding to better capture sequential information in the MDP prior, unlike the original PFN \citep{d88a5ae1673f2009704186acf2890163e6ddf4ca} that omitted it for permutation-invariant priors. However, this also increases the risk of overfitting to spurious correlations. To address this, we combine positional encoding with MAML, which helps the model generalize well by focusing on common optimization patterns rather than overfitting task-specific details. Unlike its original purpose of warm-starting parameters, MAML in our method is creatively repurposed to extract features shared across trajectories. Fine-tuning with MAML involves three steps: sampling tasks (\Lineref{line: sample agent fine tune}), computing task-specific updates (\Lineref{line: agent specific update}), and optimizing the total adapted loss (\Lineref{line: global update}). In \Secref{sec: component analysis}, we show through an ablation study that both MAML and positional encoding are crucial for building a robust trajectory surrogate.

The training cost in the first stage is incurred only once, since the base PFN can be stored and later fine-tuned for future problems. Thus, the cost for future problems is reduced to the fine-tuning stage.

\section{Experiments}\label{sec:exp}
\subsection{Experimental Setup} \label{sec: exp setups}

\paragraph{Baselines.} We compare \algname with several few-shot BO methods using different techniques: BO with meta-learned GP (\metagp), few-shot deep kernel surrogate learning (\fsbo) \citep{aea3f03299ff0cfea9b394f5559aa1c173f9876f}, pure transformer neural processes (\tnp) \citep{8ef3f06353f65ca653b396b70d089f793bfc168b, d88a5ae1673f2009704186acf2890163e6ddf4ca}, end-to-end meta-BO with transformer neural processes (\nap) \citep{maraval2023endtoend},   random search (\random), and BO with standard GP (\gp). Table~\ref{tab:techniques} shows a technical summary of them except \random and \gp since they do not use any listed techniques.

In \tnp, we meta-train a TNP \footnote{Using the same architecture as our PFN, as in \cite{maraval2023endtoend}.} on original meta-data instead of MDP priors; for comparison, we also test a variant with MAML and positional encoding, denoted as \tnpp. Similarly, we pre-train \metagp kernel parameters with meta-data before testing, and introduce a meta-trained version of our MDP prior, denoted as Meta-Acquisition Function (\maf), to align with RL-based methods. We also include \optformer \citep{943cd7cf7ba21911561d03228dc2dd3f397168c9} when possible \footnote{\optformer requires datasets with extensive optimization trajectories, which are unavailable in our drug discovery benchmarks.}. As summarized in Table~\ref{tab:techniques}, these baselines fall into two main categories: \emph{meta-surrogate learning} (\metagp, \fsbo, \tnp), which learn BO surrogates without sequential information, and \emph{meta-trajectory learning} (\maf, \optformer), which directly model trajectories. \algname and \nap use both approaches.

\begin{table}[!htbp]
    \centering
    \resizebox{0.999\textwidth}{!}{
\begin{tabular}{lccccccc}
    \toprule
    \textbf{Techniques} & \metagp & \fsbo & \tnp & \maf & \nap & \optformer & \algname (Ours) \\
    \midrule
    Meta-learn traj. & \XG & \XG & \XG & \checkmarkG & \checkmarkG & \checkmarkG & \checkmarkG \\
    Meta-learn surrog.  & \checkmarkG & \checkmarkG & \checkmarkG & \XG & \checkmarkG & \checkmarkG & \checkmarkG \\
    MAML                     & \XG & \XG & \XG & \XG & \XG & \XG & \checkmarkG \\
    Positional encoding      & \XG & \XG & \XG & \XG & \XG & \checkmarkG & \checkmarkG \\
    \bottomrule
\end{tabular}
}
    \caption{Technical summary of all compared algorithms except \random and \gp. ``\checkmarkG'' and ``\XG'' denote a certain technique is used in an algorithm or not.}
    \label{tab:techniques}
\end{table}

\paragraph{Evaluations.} 
The objective function range of each task is normalized to $[0,1]$, so the regret at iteration $t$ is defined as $1 - \max_{0\leq \tau \leq t}f(x_\tau)$ and we use the log-scaled version in our results. The rank is defined as the integer rank value of the method at a given iteration among all baselines in terms of regret performance. Both regret and rank are \textbf{the lower, the better}. As in existing literature \citep{b71e6247b14f62f9fb134e3aedff18b3cb082cb9, aea3f03299ff0cfea9b394f5559aa1c173f9876f}, for each benchmark, we report the aggregated regret and aggregated rank, computed as the average regret and rank over five different initializations, with error bars denoting 95\% confidence.

All the benchmarks contain a meta-training, meta-validation and meta-test dataset. We train the models with meta-training data. All the hyperparameters of our method (learning rate, acquisition function, MAML inner step size, fine-tuning epochs, etc.) are optimized on the validation set and the results are demonstrated on the test set. \algname uses the same pre-trained PFN for all benchmarks. We use an Adam optimizer for PFN training \citep{kingma2017adammethodstochasticoptimization}. Please refer to \Appref{app: baseline setting} for more detailed experimental settings and comparison.

\subsection{Results on Real-World Drug Discovery}
\label{sec: drug results}
The docking score estimates how strongly a small molecule (ligand) binds to a target receptor (typically a protein) where lower scores indicate stronger predicted binding affinity. In drug discovery, a favorable docking score suggests effective interaction with the disease-related receptor, potentially blocking or modulating its function. In this paper, we establish two benchmarks, each consisting of multiple molecule search tasks that aim to minimize the docking score against Covid-19 or cancer receptors, denoted as \covid and \cancer. We use a 26D continuous embedding for both datasets, converted from their \textit{mqn feature} \citep{mqn2009}. \covid and \cancer contain 5 and 2 problems respectively. Check \Appref{app: benchmark setting} for detailed dataset description. 
 
\begin{figure}[t] 
    \centering
    \begin{minipage}{\linewidth}
    \includegraphics[width=\textwidth]{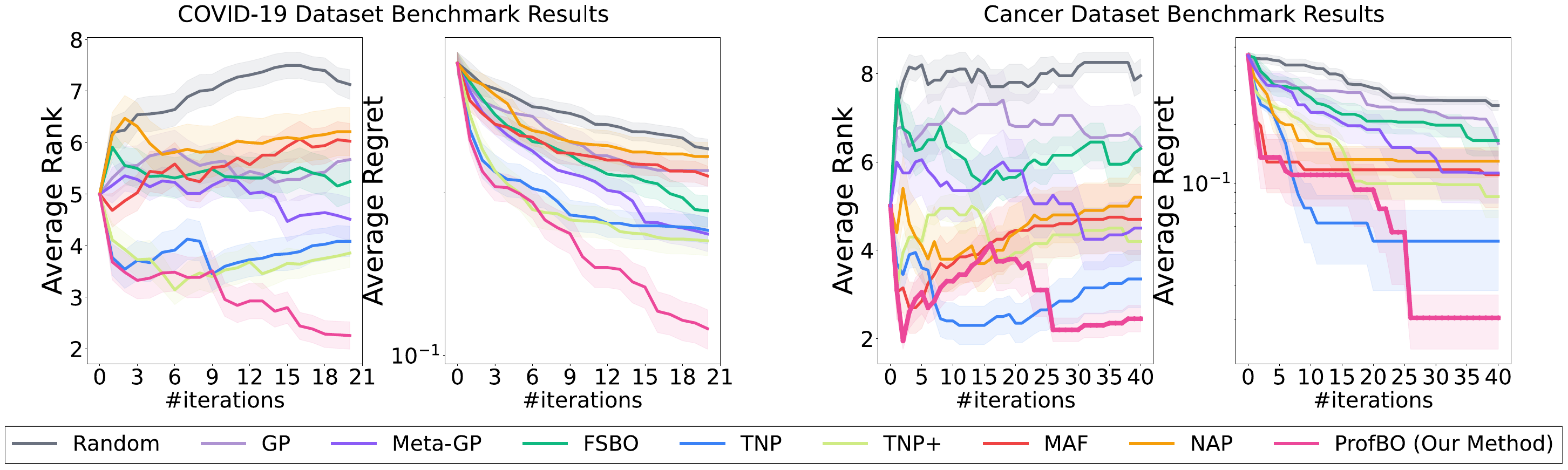}
    \end{minipage}
    \caption{Strong performance of \algname on \covid (five problems) and \cancer (two problems) compared with other baseline methods.}  
\label{fig: result covid agg}
\end{figure}

In \Figref{fig: result covid agg}, our method consistently and significantly outperforms baselines and meets the few-shot requirement by achieving strong performance within 20 or 40 iterations. In terms of average regret on both datasets, the gaps between \algname and other methods are even larger as $T$ increases. While \cite{maraval2023endtoend} reported state-of-the-art \nap’s superiority in a similar antibody design benchmark, in our experiments \nap performs comparably to the lighter \maf, likely due to the challenging high dimensionality of our benchmarks and the Transformer policy’s instability in few-shot settings. Notably, \nap required over 200 iterations to dominate in the 11D antibody design task, whereas we evaluate on 26D problems for few iterations. 

\subsection{Results on Hyperparameter Optimization}
\label{sec: hpob results}

\hpob \citep{arango2021hpoblargescalereproduciblebenchmark} is a benchmark of classification models’ hyperparameters and accuracies, widely used in few-shot BO. Following \cite{maraval2023endtoend}, we select 6 of the 16 problems, where each corresponds to the loss of a classification algorithm on different tasks, with input dimensions ranging from 2 to 18. We report results over 90 iterations to align with existing literature, though our focus is still on few-evaluation BO. Results on additional 13 problems with 25 iterations are provided in \Appref{app: additional experimental results}, showing the consistent strong performances of \algname.

\begin{figure}[!htbp] 
    \centering
    \begin{minipage}{\linewidth}
    \includegraphics[width=\textwidth]{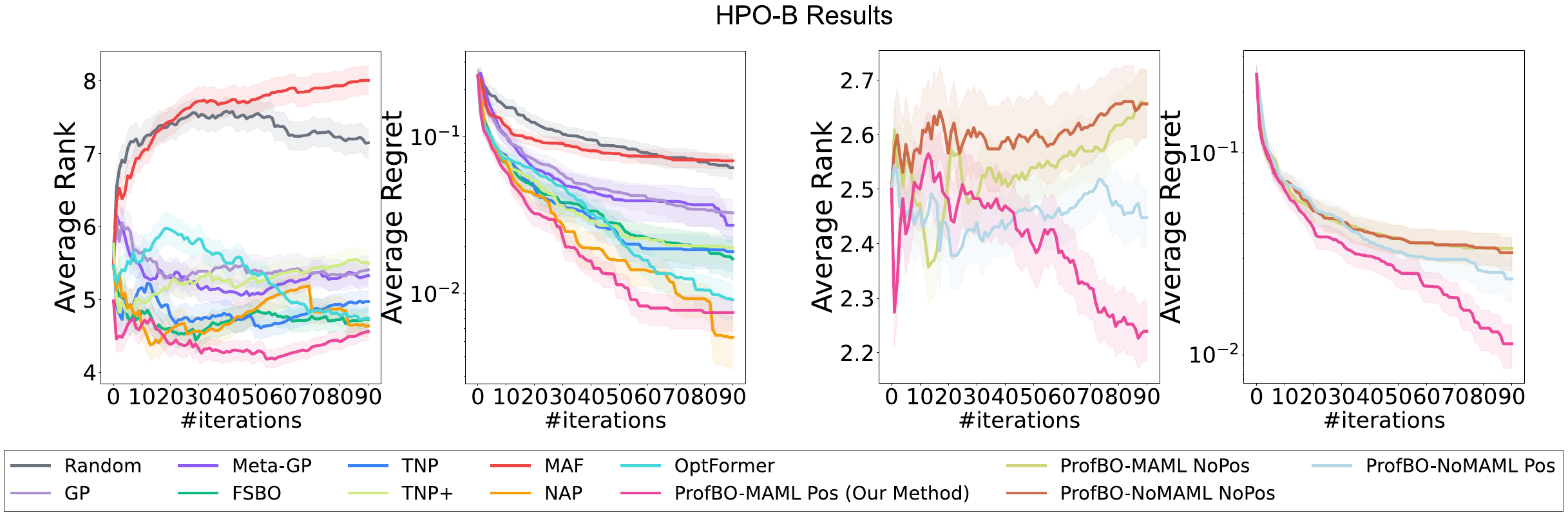}
    \end{minipage}
    \caption{[Left two] Strong performances of \algname on \hpob compared with other baseline methods.  [Right two] Ablation study results of \algname with MAML and positional encoding enabled (``MAML'', ``Pos'') or not (``NoMAML'', ``NoPos''), showing both MAML and positional encoding are important technical components of \algname. } 
\label{fig: result hpob agg}
\end{figure}

The \hpob results are shown in the left two subfigures of \Figref{fig: result hpob agg}. \algname achieves strong performances, clearly leading in average rank and regret. Meta-surrogate learning methods (\metagp, \fsbo, \tnp, \tnpp) overlook the procedural knowledge in optimization trajectories, while our surrogate leverages an MDP prior to capture this knowledge, yielding improved results. Among meta-trajectory learning methods, \maf performs even worse than \random, \optformer is competitive, and \nap, also directly modeling trajectories, outperforms both. We attribute this improvement to \nap’s supervised auxiliary loss, which helps the agent better learn inductive biases across tasks. \algname balances trajectory modeling and meta-learning by separating the two in a simple yet effective way. MAML enables efficient meta-learning of the trajectory surrogate, while PFN’s in-context learning accelerates adaptation to target tasks, yielding stronger performance than \nap and \optformer, especially in the first 10 iterations.

\subsection{Ablation Studies}
\label{sec: component analysis}

To thoroughly understand the algorithmic design of \algname, we analyze the effectiveness of three key components in the fine-tuning of \algname: the MDP prior, MAML, and positional encoding added to PFN. The test setting is the same as Section \ref{sec: hpob results}. To ensure fair comparison, for each problem, all surrogates are trained with the same dataset pre-generated from our MDP prior.

To test the effectiveness of MDP prior, we compare \algname against \tnpp. As discussed in Section \ref{sec: exp setups}, \tnpp is a meta-trained PFN using original meta-data instead of MDP priors, incorporating MAML and positional encoding, so \tnpp serves as the control group. Results shown in \Figref{fig: result covid agg}, \ref{fig: result hpob agg} provide compelling evidence for MDP priors' great contributions across three benchmarks. Without MDP priors, MAML and positional encoding provide little benefit to Transformer neural processes, as seen with \tnp and \tnpp\footnote{Consistent with \cite{aea3f03299ff0cfea9b394f5559aa1c173f9876f}, which showed that \fsbo without MAML often outperforms its MAML-enhanced variant.}. This explains why MAML is effective with MDP priors: they encode richer sequential correlations but are more prone to overfitting, where MAML helps the surrogate extract common optimization patterns. This provides evidence for our claim in \Secref{sec: MAML for traj surrogate}.

Next, to study the roles played by MAML and positional encoding, in the right two subfigures of \Figref{fig: result hpob agg}, we test four settings where MAML or positional encoding is enabled or not. The figures show that the performances of \algname indeed benefit from MAML, as the trials with MAML enabled are better than those without it under the same conditions. Also, vanilla PFN removes the positional encoding to make it permutation-invariant on the context, however, in our task we introduce the positional encoding back to PFN as it can help the surrogate better learn the optimization trajectories, which are also validated in \Figref{fig: result hpob agg}.

\subsection{Computational Efficiency}
\nap \citep{maraval2023endtoend} is one of the state-of-the-art methods in few-shot BO, and it implements a similar Transformer model to ours. \nap's efficiency was investigated in the original paper, taking only 2\% of \optformer's training time. However, \algname is even more efficient than \nap.

Both \algname and \nap use only the Transformer’s forward pass during test, resulting in similar test time, so we only compare the training time of them on the same device implementing a same Transformer architecture. \nap trains the model end-to-end, whereas \algname employs a two-stage training paradigm. We evaluate training time across five \covid problems, each with approximately 1M evaluations. \nap takes 3,925 seconds to finish the process, while \algname takes only 1,176 seconds, including 1,045 (MDP prior training) and 131 (fine-tuning) seconds, therefore, \algname is \emph{3.34 times faster} than \nap.
We attribute \algname's efficiency to its lightweight RL agent (a MLP with hidden size 200-200-200-200), which trains faster than \nap's Transformer policy. Additionally, supervised learning in \algname is less computationally intensive than RL in \nap.

\section{Conclusion} 

In many real-world black-box optimization scenarios, such as drug screening in wet labs, traditional BO methods often fail to converge within a practical number of iterations due to costly evaluations. To solve this problem, we introduce \algname, a Bayesian framework that incorporates MDP priors to transfer procedural knowledge from source tasks. Integrated with PFNs and MAML, \algname can quickly adapt to the target task in few shots. Experiments on drug discovery and hyperparameter tuning tasks demonstrate consistent improved performances of \algname over existing methods. Moreover, its modular design and efficient training process make \algname practically ready for a wide range of critical applications. Overall, this work explores the potential of applying principled Bayesian inference to the optimization trajectory prior of real-world experiments, paving the way for more efficient and generalizable optimization strategies that further accelerate scientific discovery.

\bibliography{ref}
\bibliographystyle{unsrtnat}
\appendix

\newpage
\section{Implementation Details of \algname}\label{app:profbo}
\subsection{Settings of the PFN Training}
\label{app: PFN setting}
The Transformer architecture \citep{vaswani2023attentionneed} processes sequential data by embedding each element into a vector representation and employs a specialized attention mechanism to enable elements to attend to one another during the forward pass. PFNs utilize a tailored attention mask to allow context points in $D$ to attend to each other and be attended by test locations during the forward pass. 

The implementation of the PFN is adapted from the repository of PFNs4BO \citep{b71e6247b14f62f9fb134e3aedff18b3cb082cb9} \footnote{\url{https://github.com/automl/PFNs4BO}}. Here we list the settings for the PFN used in our experiment (\Tabref{tab: pfn settings}). Similar to \cite{b71e6247b14f62f9fb134e3aedff18b3cb082cb9, d88a5ae1673f2009704186acf2890163e6ddf4ca}, our model takes contextual variables with dimension $\leq 26$ by conducting a zero-padding to the missing dimensions and normalization to the existing dimension.   
We use a discrete bar distribution for the regression head of PFN, with 1,000 bars uniformly chosen from the interval $[-4.5,4.5]$.

\begin{table}[!htbp]
    \centering
    \begin{tabular}{ll}
    \toprule
    \textbf{Hyperparameters} & \textbf{Choices}\\
    \midrule
        Embedding size & 512 \\
         Number of self-attention layers& 6\\
         Number of heads& 4\\
         Activation function & GeLU\\
         Feed forward NN size & 1024 \\
         Number of discretized bars & 1000 \\
         Pre-training learning rate& \{1e-3, 1e-4, 3e-4, 1e-5\}\\
         Fine-tuning learning rate&3e-4\\
         Batch size& 128\\
         Sequence length& 40\\
         Steps per epoch& 100\\
         Fine-tuning epoch& 2\\
         Maximum input size& 26\\
         Pre-training prior & GP\\
         Acquisition function & \{EI, PI, UCB\}\\
         Optimizer & Adam + Cosine Anneal\\
         \bottomrule
    \end{tabular}
    \caption{PFN settings \& Hyperparameters of \algname}
    \label{tab: pfn settings}
\end{table}

\subsection{Settings of the RL Training}
\label{app: RL setting}

We train our RL agent with the vanilla DQN algorithm, where there is a policy and target Q-network with identical architecture. The target network is a copy of the policy network initially, but it is updated at a specific frequency during the training process. The policy network is used to sample the trajectories, while the target network is treated as the ``ground-truth'', and is used to calculate the Bellman equation as mentioned in \Secref{sec: method}. 

In the main paper, where all baselines are discrete search problems, we developed an adapted training paradigm for the MDP prior to enable efficient learning from extensive molecule meta-data, demonstrating strong empirical performance in generating prior trajectories. Specifically, we restricted the agent to interact with only 10\% of the problem, updating to a new 10\% subset and updating the target network every 50 epochs. Initially, training multiple RL agents with a dynamic action space per epoch was challenging. In this case, increasing the action space size slowed training without significant gains. Our approach ensures robust increases in cumulative reward while substantially reducing computational costs.
Other training hyperparameters of the MDP pirior are shown in \Tabref{tab: RL settings}.  We note that the feed-forward NN of the agent has the same hidden size as the popular meta-learning neural acquisition methods \citep{hsieh2021reinforcedfewshotacquisitionfunction, 774f5e2494f037302c58b7fa549c4f1cabf7295e}. 

The sampling scheme of the final policy $\pi_\eta^{(i)}$ (i.e., $Q_\eta^{(i)}$) for task $i$ is $\epsilon$-greedy to balance exploration and exploitation:
\begin{align*}
    a_t=\left\{ 
\begin{array}{l}
\text{Uniform}(\{x^{(i)}_\tau\}_{\tau = 1}^{n_i}),\Pr =\epsilon, \\
\arg\max_{a'\in \{x^{(i)}_\tau\}_{\tau = 1}^{n_i}}, Q_\eta^{(i)}(s_t,a'),\Pr =1-\epsilon.
\end{array}
\right.
\end{align*}

\begin{table}[!htbp]
    \centering
    \begin{tabular}{ll}
    \toprule
    \textbf{Hyperparameters} & \textbf{Choices}\\
    \midrule
        Hidden size & [200, 200, 200, 200] \\
        Activation function & ReLU \\ 
        Episode length ($T$) & 40 \\
        Epoch & 250 \\
        Target update frequency & 50 \\
        Discount factor ($\gamma$) & 0.98 \\
        Maximum action sample size &$n_i//10$ \\
        History size ($k$) & 10 \\
        Replay buffer size & 10000 \\
        Learning rate & 1e-3 \\
        Optimizer & Adam \\
        \bottomrule
    \end{tabular}
    \caption{RL setting of \algname}
    \label{tab: RL settings}
\end{table}

\section{Experimental settings of other baselines}
\label{app: baseline setting}

\textbf{FSBO \footnote{\url{https://github.com/machinelearningnuremberg/FSBO}}}. \fsbo meta-trains a deep kernel GP surrogate with the meta-dataset and performs few epochs of adaptation in each BO iteration. We adopted a deep kernel GP used in \cite{aea3f03299ff0cfea9b394f5559aa1c173f9876f}. We trained the model with batch size 512 for 2000 iterations, so the total number of training data point is 1,024,000, the same as \algname. The learning rate is 1e-3.

\textbf{NAP \footnote{\url{https://github.com/huawei-noah/HEBO/tree/master/NAP}}}. We use a Transformer with the same architecture as \algname (\Tabref{tab: pfn settings}). The training completely follows the settings mentioned in the original paper, where we perform RL on a conditional GP surrogate trained with meta-data and test on discrete problem. We use a different episode length, epochs, batch size for each benchmark, so that the episode length $T$ matches the experimental results in \Secref{sec:exp} and the total training data is around 1M, please refer to our repository for details.

\textbf{OptFormer \footnote{\url{https://github.com/google-research/optformer}}}. The results in \hpob (\Secref{sec: hpob results}) is produced by the authors of \nap based on the official codebase. Please refer to their paper for details.

\textbf{TNP.} We used the same PFN architecture with a positional encoding as in \Tabref{tab: pfn settings}. As mentioned in \Secref{sec: component analysis}, we used 1M raw samples from meta-dataset to train the surrogate model with MAML and positional encoding.

\textbf{Meta-GP.} Like \cite{maraval2023endtoend}, we also pre-train the RBF kernel parameters of the GP with the meta-dataset and initialize the GP with pre-trained parameters at the test time. The implementation is based on BO package \textsc{BoTorch} \citep{NEURIPS2020_f5b1b89d}.  

\textbf{NAF.} Please refer to \Tabref{tab: RL settings} for the settings of the RL agent. We used a RL prior that learns from all the mata-datasets, instead of a single dataset. To do that, we train the RL agent to optimize $\mathcal{D}^{(i)}, i \in [N]$ for many $i$. Specifically, we sample $i$ uniformly from $[N]$ in each outer loop and train the agent in the same way as described in \Appref{app: RL setting}. 

\section{Experimental Details of Benchmarks}
\label{app: benchmark setting}

\subsection{HPO-B}

Please refer to official repository \footnote{\url{https://github.com/machinelearningnuremberg/HPO-B}} and paper \cite{arango2021hpoblargescalereproduciblebenchmark} for detailed explanation of the dataset. The six representative problems in \Secref{sec: hpob results} are 5860 (glmnet), 4796 (rpart.preproc), 5906 (xgboost), 5889 (ranger), 5859 (rpart), 5527 (svm).
We also provide the aggregated results of 13 \hpob problems in the following section of appendix (\Figref{fig: hpob 13 sub-problems}).
\subsection{Covid-B and Cancer-B}

The \covid dataset comprises five problems as in \cite{arango2021hpoblargescalereproduciblebenchmark} across 24 tasks. Each task within a problem optimizes the docking score for a shared Covid-19 receptor but targets different binding interactions. For instance, task \texttt{NPRBD\_6VYO\_AB\_1\_F} involves interactions with both chains A and B of the PDB structure 6VYO, while \texttt{NPRBD\_6VYO\_A\_1\_F} focuses solely on chain A. The validation set consists of unique datasets from \cite{liu2023drugimprover} that are distinct from their parent structure.
The \cancer dataset differs from \covid, as it lacks related tasks from a common parent structure. Instead, it includes docking scores for 5 distinct cancer-target receptors, manually divided into source and target tasks. The validation set is a mixed random sample drawn from each dataset. The final dataset comprises three training sets and two test sets. 

For both datasets, the problem for each task is a set of molecules, with size ranging from 2K to 150K and are stored as Simplified Molecular Input Line Entry System  (SMILES). We first convert the molecules to their \textit{mqn feature} (42D, integer) \citep{mqn2009}, then we use principle component analysis (PCA) to construct a 26D continuous representation for each molecule, where the percentage of explained variance in PCA are greater than $95\%$ for all problems.

\textbf{Covid-B.} The \covid dataset is adapted from the Covid dataset used in DrugImprover \citep{liu2023drugimprover} \footnote{\url{https://github.com/xuefeng-cs/DrugImprover}}. They were chosen by the authors from the Zinc 15 dataset \citep{zinc15}, containing 11M drug-like molecules. The original dataset consists of 24 \texttt{.csv} files, each containing 1,000,000 molecular docking samples across various SARS-CoV-2 receptors. We identified five structural conditions shared by multiple receptors, each defined as a single \textit{problem} (in \hpob, one learning algorithm corresponds to one problem). While different \textit{datasets} in \hpob share the same supervised loss but differ in training data, in \covid datasets correspond to the same structural condition under varying sub-conditions. Finally, we retain only the molecules common across their respective problems in each dataset. All the objective values are normalzied to $[0,1]$. 

\begin{table}[!htbp]
    \centering
    \begin{tabular}{lcccc}
    \toprule
        Problems & \multicolumn{2}{c}{Meta-training} & \multicolumn{2}{c}{Meta-test} \\
        \cmidrule(lr){2-3} \cmidrule(lr){4-5}
         & \# Evaluations & \# Datasets & \# Evaluations & \# Datasets \\
        \noalign{\smallskip} \hline \noalign{\smallskip}
        \texttt{NPRBD} &  46,176&3 &  30,784& 2 \\
        \texttt{NSP10-16} & 202,254 & 1 & 202,254 & 1 \\
        \texttt{NSP15} & 11,540 &  5& 4,616 & 2 \\
        \texttt{Nsp13.helicase} & 244,361 & 1 & 244,361 &  1\\
        \texttt{RDRP} & 106,266 & 3 & 35,422 & 1 \\
        \bottomrule
    \end{tabular}
    \caption{Statistics of each \covid problem in the meta-training and meta-test datasets. }
    \label{tab: covid dataset}
\end{table}

We provide a list of problems of \covid and the size of each meta-train and meta-test dataset in \Tabref{tab: covid dataset}. All the problems use a common meta-validation dataset, containing 4 datasets chosen as subsets of \texttt{3CLPro\_7BQY\_A\_1\_F, NSP16\_6W61\_A\_1\_H, PLPro\_6W9C\_A\_2\_F, NSP10\_6W61\_B\_1\_F} and 1M evaluations in total. 

\textbf{Cancer-B.}
We also refer to \cite{liu2023drugimprover, zinc15} for the original five sets of molecules docked on different cancer proteins.
In \cancer, we utilize three meta-training datasets (\texttt{6T2W}, \texttt{NSUN2}, \texttt{RTCB}), comprising 437,634 evaluations in total, and two meta-test datasets (\texttt{WHSC}, \texttt{WRN}), totaling 291,756 evaluations. The meta-validation set is constructed as a balanced mixture of random samples from these five datasets, totaling to 10,000 evaluations, with each dataset contributing an equal proportion. Molecules are also common in five \texttt{.csv} files.

\section{Additional experimental results}
\label{app: additional experimental results}
In this section, we present the additional experimental results.

\label{sec: inference time compare}

\subsection{Additional Baseline Comparison}
\label{sec: additional figures}
In the following, we demonstrate the baseline comparison of regret and rank in each problem, i.e., the results in \Secref{sec: drug results}, \ref{sec: hpob results} before aggregation w.r.t. problems.

\begin{figure}[!htbp]
    \centering
    \begin{subfigure}{\linewidth}
        \centering
        \includegraphics[width=\textwidth]{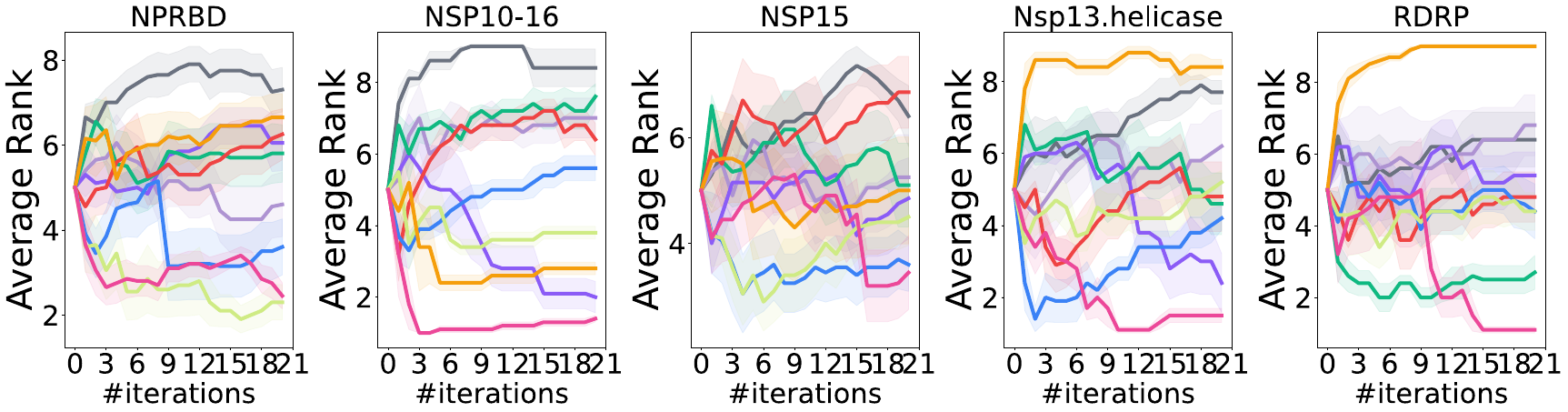}
        \caption{Rank of each problem of \covid}
        \label{subfig:covid_rank}
    \end{subfigure}
    \vspace{2mm} 
    \begin{subfigure}{\linewidth}
        \centering
        \includegraphics[width=\textwidth]{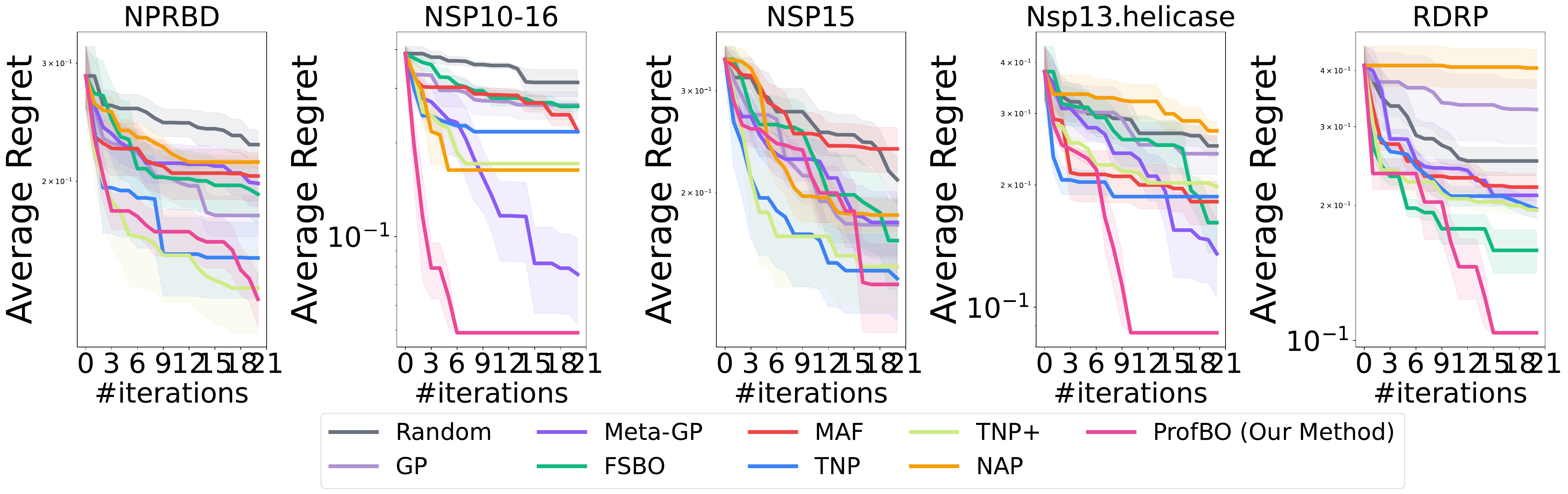}
        \caption{Regret of each problem of \covid}
        \label{subfig:covid_regret}
    \end{subfigure}
    \caption{Performance comparisons on \covid.}
    \label{fig: covid more}
\end{figure}

\begin{figure}[!htbp]
    \centering
    \begin{subfigure}{0.49\linewidth}
        \centering
        \includegraphics[width=\textwidth]{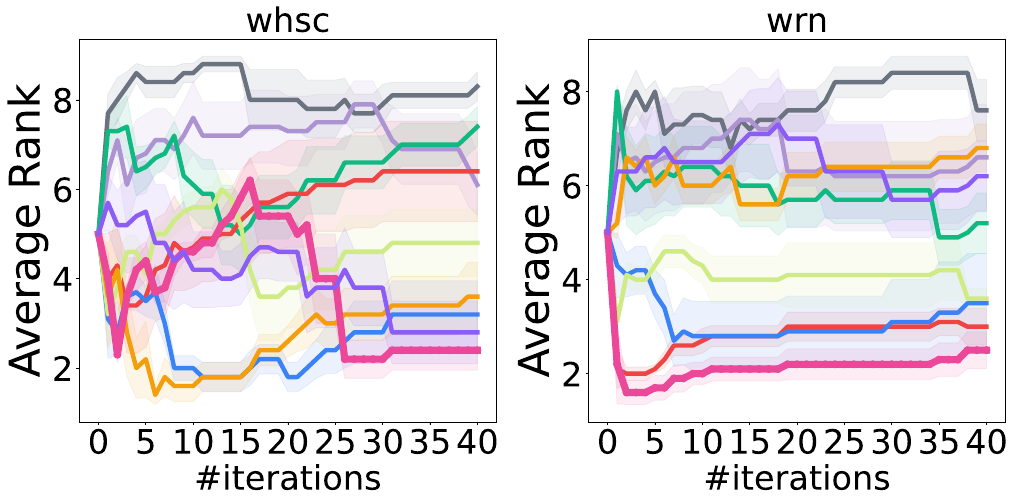}
        \caption{Rank of each problem of \cancer}
        \label{subfig:cancer_rank}
    \end{subfigure}
    \vspace{2mm} 
    \begin{subfigure}{0.49\linewidth}
        \centering
        \includegraphics[width=\textwidth]{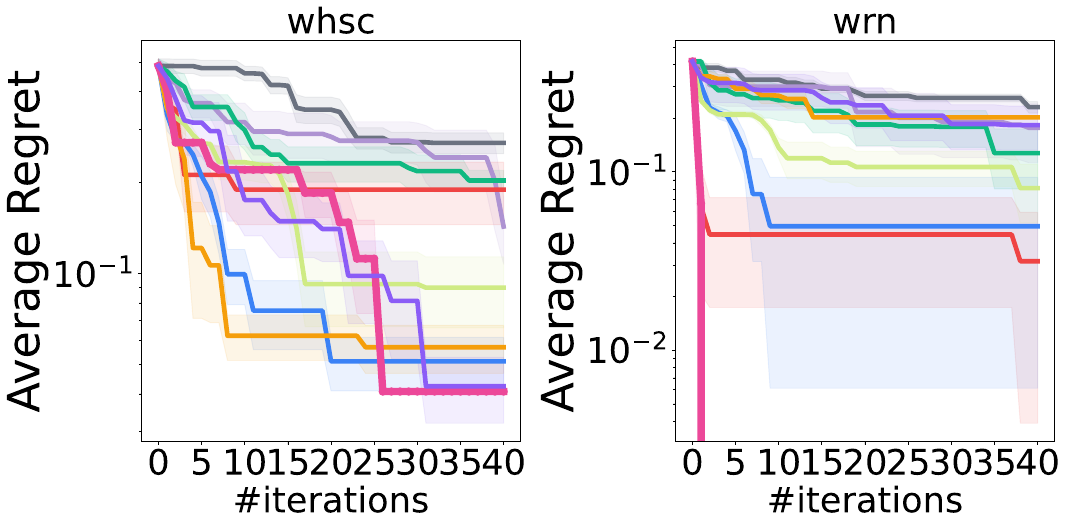}
        \caption{Regret of each problem of \cancer}
        \label{subfig:cancer_regret}
    \end{subfigure}
\begin{subfigure}{0.99\linewidth}
        \centering
        \includegraphics[width=\textwidth]{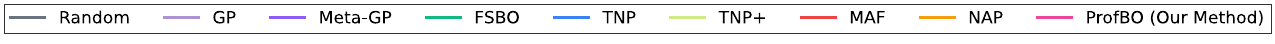}
    \end{subfigure}
    \caption{Performance comparisons on \cancer.}
    \label{fig: cancer more}
\end{figure}

\Figref{fig: covid more} demonstrates the results on 5 problems in \covid, where \algname shows superior performance in general, while \nap is the most unstable one.
\Figref{fig: cancer more} demonstrates the results on 2 problems in \cancer, where \algname demonstrates robust performances by being the best method after 30 iterations. Other methods suffer from instability in different problems.

\begin{figure}[!htbp]
    \centering
    \begin{subfigure}{0.95\linewidth}
        \centering
        \includegraphics[width=0.95\textwidth]{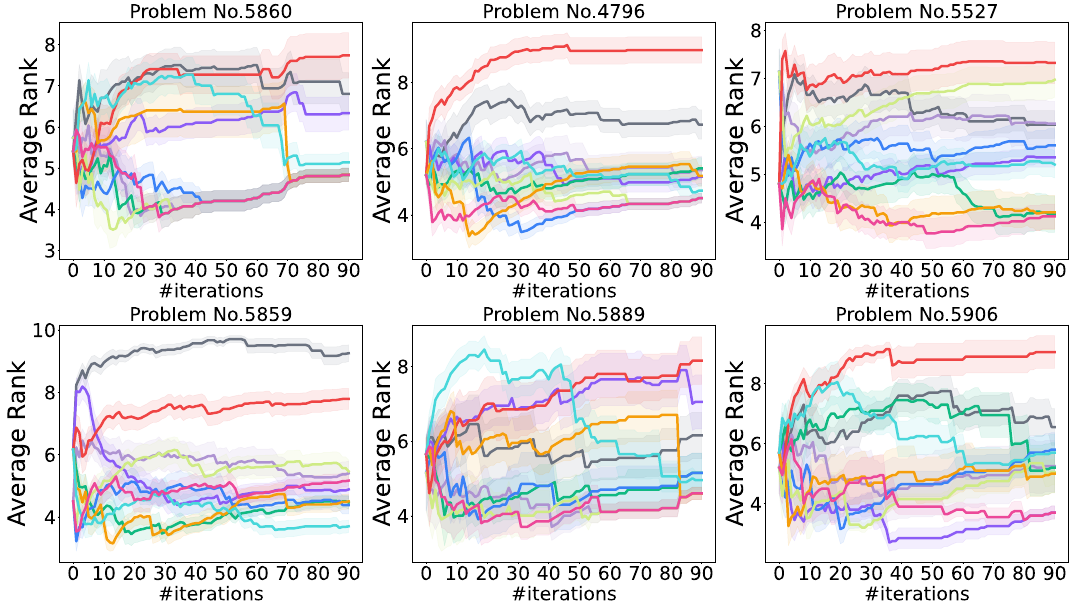}
        \caption{Rank of each problem of \hpob}
        \label{subfig:hpob_rank}
    \end{subfigure}
    \vspace{2mm} 
    \begin{subfigure}{0.95\linewidth}
        \centering
        \includegraphics[width=0.95\textwidth]{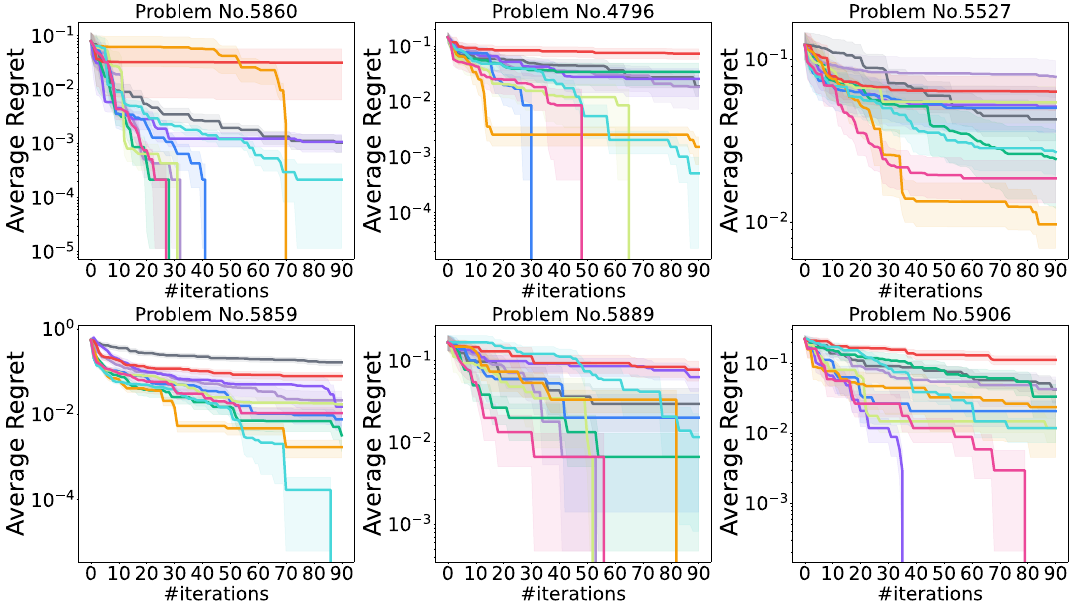}
        \caption{Regret of each problem of \hpob}
        \label{subfig:hpob_regret}
    \end{subfigure}

    \begin{subfigure}{0.85\linewidth}
        \centering
        \includegraphics[width=0.85\textwidth]{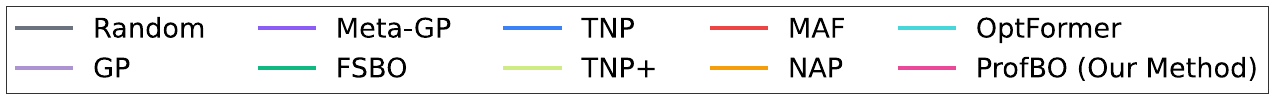}
    \end{subfigure}
    \caption{Performance comparisons on \hpob.}
    \label{fig: hpob more}
\end{figure}

\begin{figure}[!htbp]
    \centering
\includegraphics[width=0.6\textwidth]{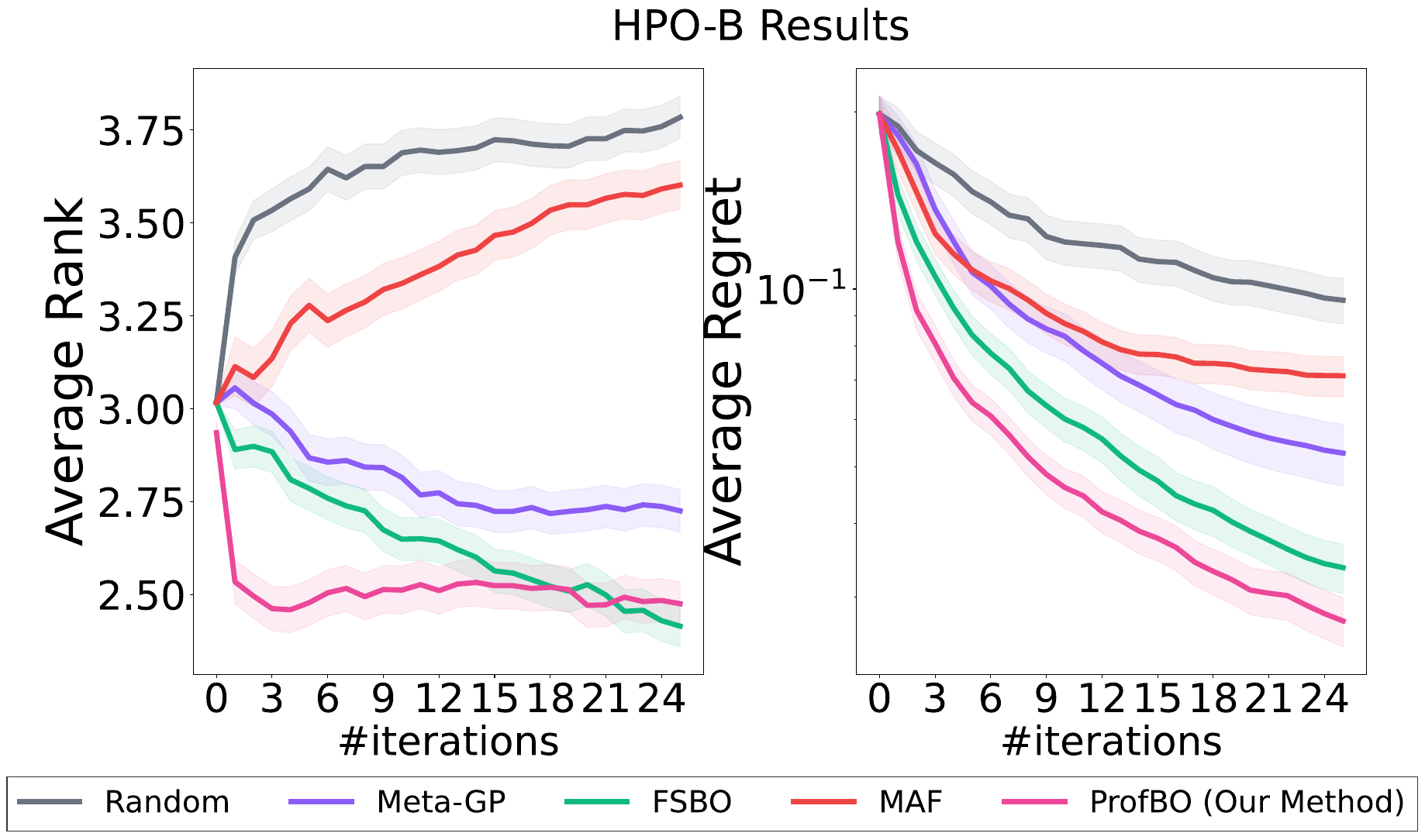}
    \caption{Aggregated results of \hpob with 13 problems for 25 iterations.}
    \label{fig: hpob 13 sub-problems}
\end{figure}

\Figref{fig: hpob more} shows the results on 6 \hpob problems. We can see that the baselines' performance stabilizes after 90 iterations, and \algname is consistently in the top 3 baselines except for Problem No. 5859. This shows that \algname is also prominent in problems that require long iterations. In \Figref{fig: hpob 13 sub-problems}, we demonstrate the aggregated results of \algname on 13 problems of \hpob for 25 iterations. We find \algname still excels in these problems, providing more solid evidence of its ability to adapt to problems with varying input dimensions and diverse meta-task dependencies.

\subsection{Additional Function Prediction Comparison}
\begin{figure}[!htbp]
    \centering
\includegraphics[width=0.99\textwidth]{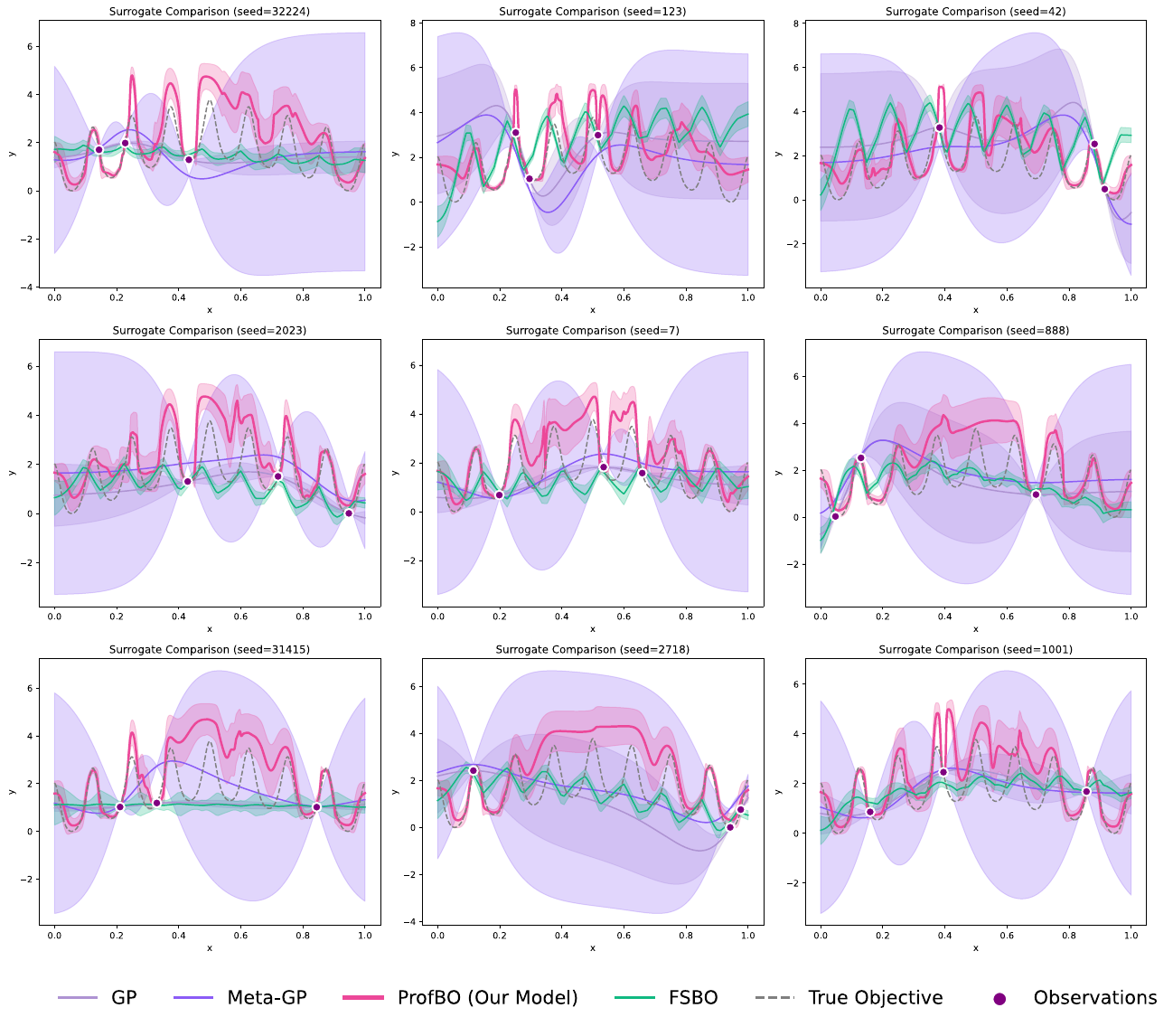}
    \caption{Function prediction comparison.}
    \label{fig: surrogate comparison more}
\end{figure}

In \Figref{fig: surrogate comparison more}, we demonstrate more visualizable comparison between \algname, \fsbo, and \gp, similar to \Figref{fig: surrogate comparison}. In addition, we add \metagp to our comparison where the GP parameters are pretrained with meta-datasets. The settings for \algname, \fsbo and \gp are the same as mentioned in \Figref{fig: surrogate comparison}. As we can see, \algname consistently models the true objective function much better than other methods, with only 3 evaluation points.

\end{document}